\documentclass[10pt,twocolumn,letterpaper]{article}

\usepackage{iccv}
\usepackage{times}
\usepackage{epsfig}
\usepackage{graphicx}
\usepackage{amsmath}
\usepackage{amssymb}
\usepackage{subfigure}
\usepackage{booktabs} 
\usepackage{multirow}
\usepackage{ulem}
\usepackage{tabularx}
\usepackage{tabu}

\usepackage{algorithm}
\usepackage{algorithmic}

\usepackage[breaklinks=true,bookmarks=false]{hyperref}

\iccvfinalcopy 

\def\iccvPaperID{****} 

\ificcvfinal\fi

\begin{document}

\title{Removing Adversarial Noise in  Class Activation Feature Space}

\author{Dawei Zhou\\
Xidian University\\
{\tt \small dwzhou.xidian@gmail.com}
\and
Nannan Wang\\
Xidian University\\
{\tt \small nnwang@xidian.edu.cn}
\and
Chunlei Peng\\
Xidian University\\
{\tt \small clpeng@xidian.edu.cn}
\and
Xinbo Gao\\
Chongqing University of\\Posts and Telecommunications\\
{\tt \small gaoxb@cqupt.edu.cn}
\and
Xiaoyu Wang\\
The Chinese University of\\Hong Kong (Shenzhen)\\
{\tt \small fanghuaxue@gmail.com}
\and 
Jun Yu\\
University of Science and\\Technology of China\\
{\tt \small  harryjun@ustc.edu.cn}
\and
Tongliang Liu\\
The University of Sydney\\
{\tt\small tongliang.liu@sydney.edu.au}
}

\maketitle
\ificcvfinal\thispagestyle{empty}\fi

\begin{abstract}
Deep neural networks (DNNs) are vulnerable to adversarial noise. Preprocessing based defenses could largely remove adversarial noise by processing inputs. However, they are typically affected by the \textit{error amplification effect}, especially in the front of continuously evolving attacks. To solve this problem, in this paper, we propose to remove adversarial noise by implementing a self-supervised adversarial training mechanism in a class activation feature space. To be specific, we first maximize the disruptions to class activation features of natural examples to craft adversarial examples. Then, we train a denoising model to minimize the distances between the adversarial examples and the natural examples in the class activation feature space. Empirical evaluations demonstrate that our method could significantly enhance adversarial robustness in comparison to previous state-of-the-art approaches, especially against \textit{unseen} adversarial attacks and \textit{adaptive} attacks.
\end{abstract}

\section{Introduction}
\label{section1}
Deep neural networks (DNNs) are known to be vulnerable to adversarial examples. Adversarial examples are maliciously crafted by adding imperceptible but adversarial noise on natural examples \cite{goodfellow2014explaining,szegedy2013intriguing,shen2017ape,liao2018defense,ma2018characterizing,wu2020adversarial}. The vulnerability of DNNs poses a potential threat to many decision-critical deep learning applications, such as image processing \cite{lecun1998gradient,he2016deep,Zagoruyko2016WRN,simonyan2014very,2017Mask,ma2021understanding} and natural language processing \cite{sutskever2014sequence}. Thus, it is important to find an effective defense against adversarial noise.

\begin{figure}[t]
\vskip 0.2in
\begin{center}
\centerline{\includegraphics[width=3.2 in]{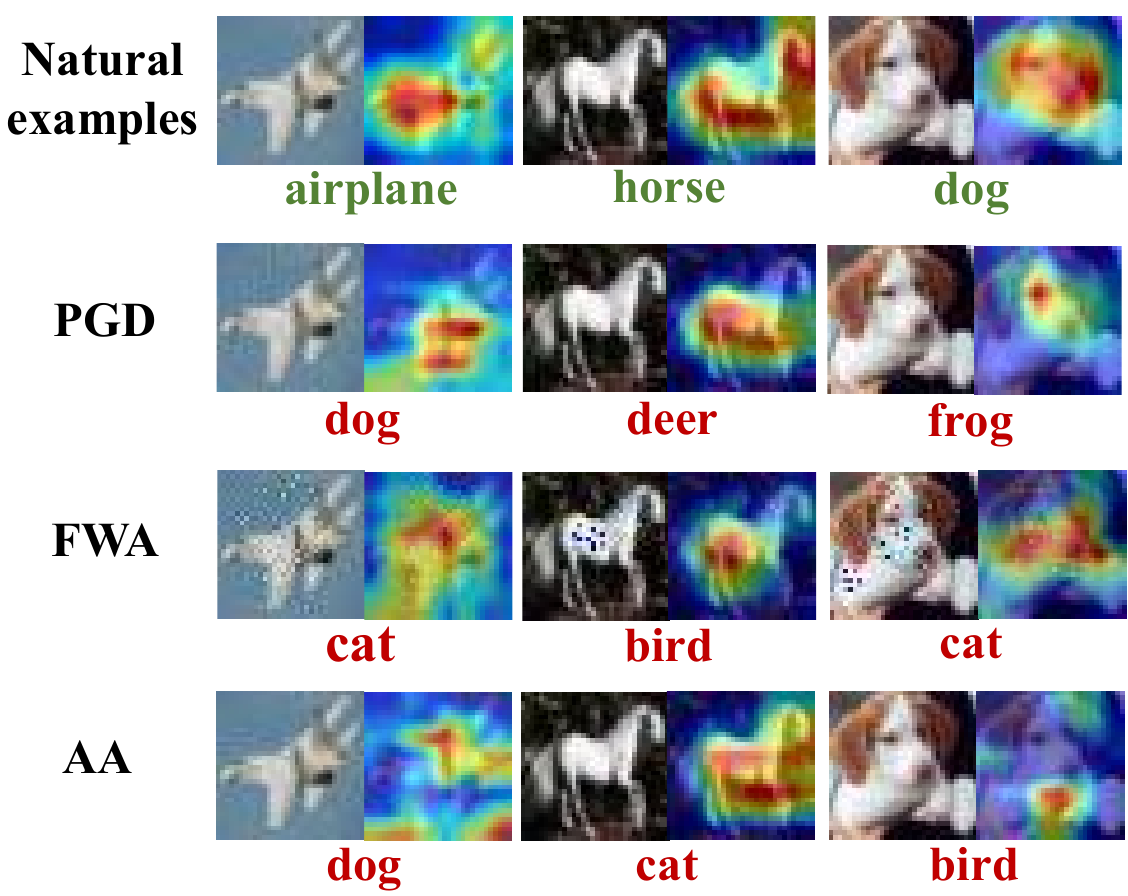}}
\caption{A visual illustration of class activation maps of natural examples and adversarial examples. The adversarial examples are crafted by distinct types of non-targeted attacks, i.e., PGD \cite{madry2017towards}, FWA \cite{wu2020stronger} and AA \cite{croce2020reliable}. Although adversarial noise is imperceptible in pixel level, there exists obvious discrepancies between the class activation maps of natural examples and adversarial examples.}
\label{fig1}
\end{center}
\vskip -0.4in
\end{figure}

Previous researches show that adversarial robustness of target models could be enhanced by processing inputs with certain transformations to reduce noise \cite{gu2014towards,das2017keeping,meng2017magnet,guo2017countering,liao2018defense}. However, preprocessing based defenses may suffer from the \textit{error amplification effect}, in which small residual adversarial noise is amplified to a large perturbation in internal layers of the target model and leads to misleading predictions \cite{liao2018defense}. Furthermore, these preprocessing based approaches are shown to be less effective in front of unseen adversarial attacks \cite{wu2020stronger,croce2020reliable,xiao2018spatially} as the adversarial perturbations of adversarial examples they used may not be the maximum in internal layers (see Section~\ref{section4}).

The \textit{class activation mapping} technique \cite{zhou2016learning} gives us an inspiration to solve this problem. Given a classification network, the class activation mapping technique could identify the importance of image regions by projecting back the class weights of the output layer on to the last convolutional features and performing a linear sum of the weighted features \cite{zhou2016learning}. We find that although adversarial noise is imperceptible in pixel level, there exists obvious discrepancies between the class activation maps of natural examples and adversarial examples crafted by existing attack approaches (see Figure~\ref{fig1}). In addition, the weighted features are in the high-level layer of the network where small residual noise could cause large perturbations. This motivates us to handle the issue of error amplification effect by designing a defense method which focuses on the weighted features called as \textit{class activation features}.

In this paper, we propose an adversarial training mechanism to remove adversarial noise by exploiting class activation features. In a high level, we design a max-min formula in the class activation feature space to learn a denoising model in a self-supervised manner without seen types of adversarial examples and ground-truth labels. Specifically, we first craft adversarial examples by maximally disrupting the class activation features of natural examples. The discrepancies of class activation features make adversarial examples have different prediction results from natural examples. We name such attack as \textit{class activation feature based
attack} (CAFA). Then, we train a denoising model, namely \textit{class activation feature based denoiser} (CAFD), to remove adversarial noise. Instead of directly utilizing pixel-level loss functions to train our model, we minimize the distances between the class activation features of the natural examples and the adversarial examples. Finally, an image discriminator is introduced to make restored examples close to the natural examples by enhancing the fine texture details. 

Achieved by such self-supervised adversarial training, our defense method could provide more significant protections against unseen types of attacks and adaptive attacks compared to previous defenses, which is empirically verified in Section~\ref{section4.2}. Furthermore, additional evaluations on ablation study and robustness of our model to the perturbation budget in Section~\ref{section4.3} further demonstrate the effectiveness of our method. 

The main contributions in this paper are as follows:
\begin{itemize}
    \item We find that although adversarial noise is imperceptible in pixel level, it significantly disrupts the class activation features of natural examples. To this end, we design a \textit{class activation features based denoiser} (CAFD) to effectively remove adversarial noise by exploiting class activation features.
    \item An self-supervised adversarial training mechanism is proposed to train the denoiser. We maximally disrupting the class activation features of natural examples to craft adversarial examples, and use them to train the denoiser for learning to minimize the distances between natural and adversarial examples in the class activation feature space.
    
    
    \item Empirical experiments show that our method could enhance adversarial robustness and it could be transferred across different target models. Particularly, the success rates of unseen attacks and adaptive attacks are reduced significantly in comparison to previous state-of-the-art approaches.
\end{itemize}

The rest of this paper is organized as follows. In Section~\ref{section2}, we briefly review related work on attacks and defenses. In Section~\ref{section3}, we describe our defense method and present its implementation. Experimental results on different datasets are provided in Section~\ref{section4}. Finally, we conclude this paper in Section~\ref{section5}.

\section{Related work}
\label{section2}
\textbf{Attacks:} Adversarial examples have been shown to be able to mislead DNNs \cite{szegedy2013intriguing} and transfer across different target models \cite{liu2016delving}. They can be crafted by single-step or multi-step attacks following the direction of adversarial gradients under a $L_{p}$ norm perturbation budget. Attacks based on this strategy include fast gradient sign method (FGSM) \cite{goodfellow2014explaining}, basic iterative attack (BIA) \cite{kurakin2016adversarial}, the strongest first-order information based projected gradient descent (PGD) method \cite{madry2017towards}, Carlini and Wagner (CW) method \cite{carlini2017towards}, decoupling direction and norm (DDN) method \cite{rony2019decoupling} and the autoattack (AA) method \cite{croce2020reliable}. Rather than optimizing the objective function at a single point, the translation-invariance input diversity method (TI-DIM) \cite{dong2019evading,xie2019improving} uses a set of translated images to optimize an adversarial example. In addition, unlike these pixel-constrained attacks which do not consider the semantic or geometric information, spatially-constrained attacks focus on mimicking non-suspicious vandalism via geometry and spatial transformation, e.g., faster wasserstein attack (FWA) \cite{wu2020stronger} and spatial transform attack (STA) \cite{xiao2018spatially}.

\begin{figure*}[t]
\begin{center}
\centerline{\includegraphics[width= 6.5in]{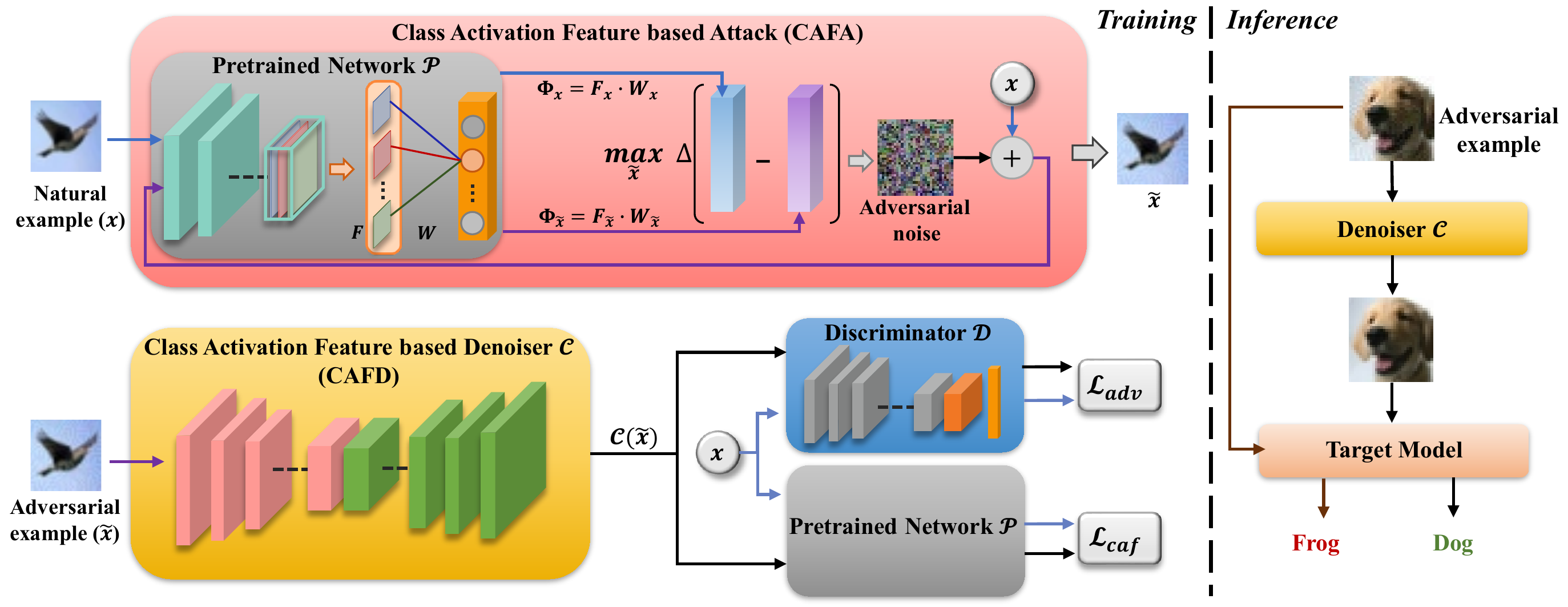}}
\caption{A visual illustration of our defense method. The proposed defense learns to remove adversarial noise via a self-supervised adversarial training mechanism. We maximally disrupt the class activation features of natural examples to craft adversarial examples and use them to train the denoiser for learning to bring adversarial examples close to natural examples in the class activation feature space.}
\label{fig2}
\end{center}
\vskip -0.4in
\end{figure*}

\textbf{Defenses:} Adversarial training (AT) is an extensive strategy for defending adversarial noise \cite{goodfellow2014explaining,wang2019improving,kurakin2016adversarial}. It is dedicated to train a robust model by augmenting the training data with adversarial examples \cite{madry2017towards}. For example, defensing against occlusion attacks (DOA) \cite{wu2019defending} method use adversarial examples crafted by occlusion attacks to enhance the backbone model's robustness. Channel-wise Activation Suppressing (CAS) \cite{bai2021improving} strategy suppresses redundant activations from being activated by adversarial perturbations during the adversarial training process. Adversarial training could improve the accuracy of the target model on adversarial examples, but it typically cannot directly be transferred to other models or tasks.

Preprocessing based methods process the inputs to achieve robustness against adversarial noise. For example, JPEG compression \cite{guo2017countering} and total variation minimization (TVM) \cite{guo2017countering} were proposed to remove high-frequency components and small localized changes respectively. Jin \textit{et al.} \cite{shen2017ape} proposed APE-G to back adversarial examples close to natural examples via a generative adversarial network. A high-level representation guided denoiser (HGD) \cite{liao2018defense} method was utilized as a pre-processing step to remove adversarial noise. Different from the above methods, we combine the benefits of adversarial training and preprocessing
and design an denoising model that remove adversarial noise by minimizing the distances between the class activation features of natural and adversarial examples in a self-supervised adversarial training manner.

\section{Methodology}
\label{section3}
\subsection{Preliminaries}
\label{section3_1}
In this paper, we aim to design a preprocessing based defense which could mitigate the error amplification effect and provide robust protections. The basic intuition behind our defense is to effectively exploit the class activation features of DNNs which could be significantly disrupted by adversarial noise. Towards this end, we design a \textit{class activation features based denoiser} (CAFD) which learns to remove adversarial noise in the class activation feature space. To train our CAFD, we propose a self-supervised adversarial training mechanism without using seen types of adversarial examples and ground-truth labels. As shown in Figure~\ref{fig2}, the training procedure can be regarded as a max-min formula and it is expressed as in the following:

For a given natural example $x$, let $\Phi_x$ represent its class activation features obtained from a pretrained deep neural network $\mathcal{P}$. We first craft an adversarial example $\tilde{x}$ by maximally disrupting $\Phi_x$ (Section~\ref{section3_2}). Then, at the minimization step, a class activation feature based denoiser $\mathcal{C}$ tries to remove the adversarial noise by minimizing the discrepancies between $\Phi_x$ and $\Phi_{\tilde{x}}$, where $\Phi_{\tilde{x}}$ denotes the class activation feature of $\tilde{x}$. In addition, to further enhance fine texture details of restored examples, we introduce an image discriminator $\mathcal{D}$ to play a game with $\mathcal{C}$ (Section~\ref{section3_3}). 

\subsection{Crafting adversarial examples}
\label{section3_2}
Our defense model is trained in an adversarial manner without utilizing seen types of adversarial examples and ground-truth labels. The adversarial examples used for self-supervised training are obtained by \textit{class activation feature based attack} (CAFA). Below, we first outline class activation features and the impact of disrupting class activation features. Then, we describe the procedure of CAFA.

\vspace{0.5em}\noindent\textbf{Class activation features:}
Given a pretrained deep neural network $\mathcal{P}$, the class activation mapping technique \cite{zhou2016learning} projects back class weights of the output layer of $\mathcal{P}$ on to the last convolutional features and performs a linear sum of the weighted features. To be specific, for a given example $x$, its predicted probability of class $c$ is $p(c|x)$. $c_{x}=\arg \max _{c} p(c|x)$ is the predicted class of $x$. We first use $f_{x}^{k}$ to represent the deep feature of $x$ of the $k-th$ channel in the last convolutional layer of $\mathcal{P}$. Then, for class $c_{x}$, the weighted feature of the $k-th$ channel is $\phi_{x}^{k}= f_{x}^{k} \cdot w_{x}^{k}$, where $w_{x}^{k}$ is the class weight of the $k-th$ channel corresponding to class $c_{x}$. Essentially, $w_{x}^{k}$ indicates the importance of $f_{x}^{k}$ for $c_{x}$ \cite{zhou2016learning}. By linear summation of all $\phi_{x}^{k}$, we could get a class activation map of $x$. We name the weighted features for all $K$ channels as \textit{class activation features} which are denoted by $\Phi_{x} = \left[\phi_{x}^{1},\phi_{x}^{2}, \ldots, \phi_{x}^{K}\right]^{\top}$. Intuitively, the class activation features could be expressed as $\Phi_{x}=F_{x} \cdot W_{x}$, where $F_{x}$=$\left[f_{x}^{1},f_{x}^{2}, \ldots, f_{x}^{K}\right]^{\top}$ and $W_{x}$=$\left[w_{x}^{1},w_{x}^{2}, \ldots, w_{x}^{K}\right]^{\top}$ are the deep features and class weights for all $K$ channels respectively.


\vspace{0.5em}\noindent\textbf{Disruptions to class activation features:}
We note that there exists obvious discrepancies between class activation maps of natural examples and adversarial examples crafted by existing attacks. Since class activation maps is the liner sum of class activation features, the discrepancies indicate that adversarial noise could significantly disrupt the class activation features. This is similar to the phenomenon described in the error amplification effect that residual adversarial noise could cause large perturbations in internal layers of a target model. The reason why we use class activation features to craft adversarial examples is that the disruptions to class activation features could directly impact the effect of misleading the target model. To show this, we conduct a proof-of-concept experiment. 

\begin{figure}[t]
\begin{center}
\centerline{\includegraphics[width=3.2 in]{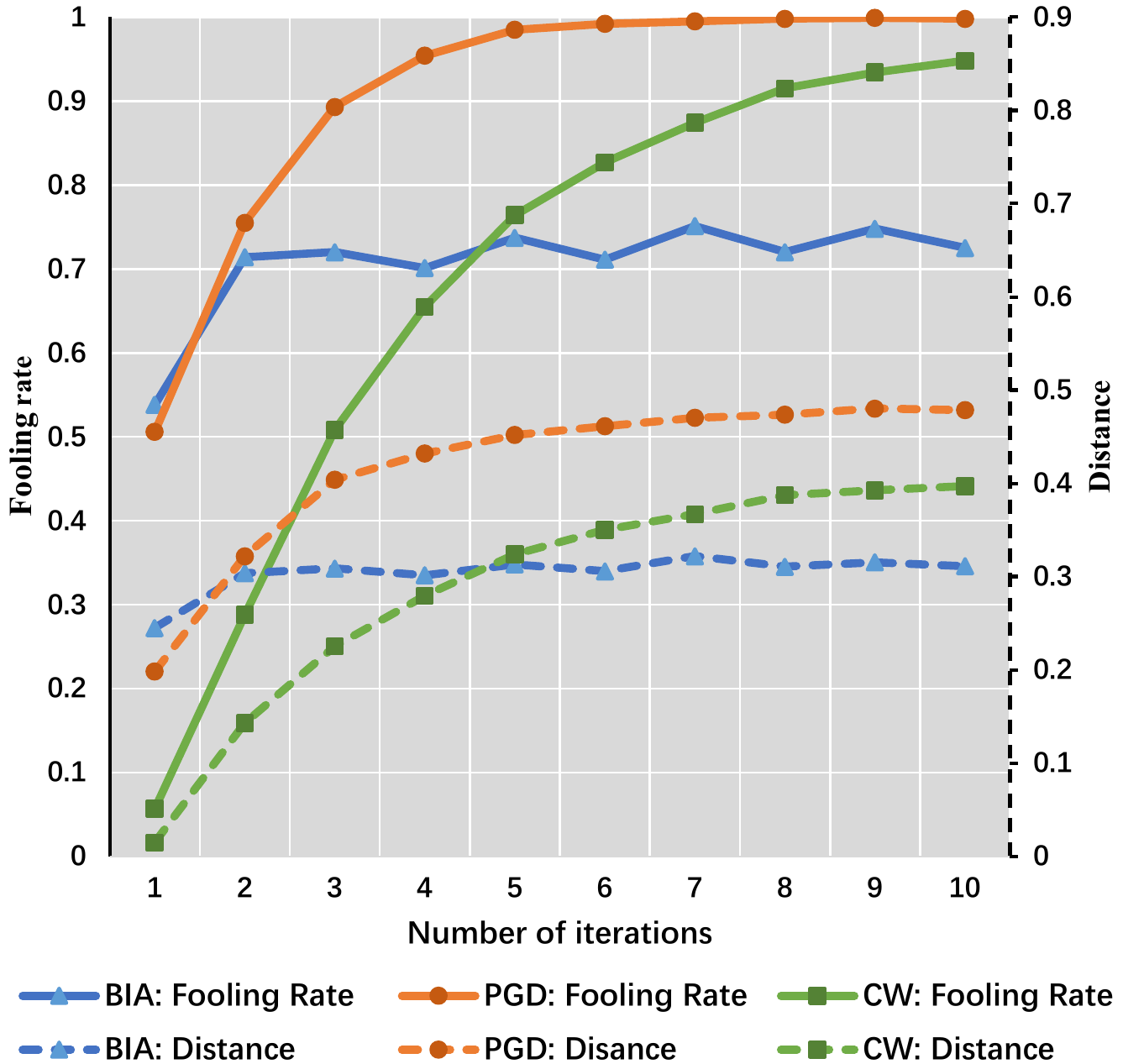}}
\caption{The average feature distances and fooling rate of adversarial examples against a VGG-19 \cite{simonyan2014very} target model on \textit{CIFAR-10} \cite{krizhevsky2009learning}. The adversarial examples are respectively crafted by BIA\cite{kurakin2016adversarial}, PGD \cite{madry2017towards} and CW \cite{carlini2017towards}. It could be seen from the figure that the fooling rates rise synchronously as the feature distances increase.}
\label{fig3}
\end{center}
\vskip -0.4in
\end{figure}

We define the \textit{feature distance} to measure the disruptions to the class activation features of a natural example $x$ as follows:
\begin{equation}
\label{eq1}
\Delta(x, \tilde{x})=\delta(\Phi_{x}, \Phi_{\tilde{x}}) \text{,}
\end{equation}
where $\Phi_{x}$ and $\Phi_{\tilde{x}}$ are the class activation features of $x$ and its adversarial example $\tilde{x}$ respectively. $\delta(\cdot)$ denotes a $L_2$-norm distance metric. As shown in Figure~\ref{fig3}, we implement three classic attack methods, BIA \cite{kurakin2016adversarial}, PGD \cite{madry2017towards} and CW \cite{carlini2017towards}. The results show that the fooling rates and the feature distances have the same changing trend. This indicates that the disruptions to class activation features could directly impact the attack effect. Therefore, maximizing the objective in Eq.~\ref{eq1} could craft strong adversarial examples and enable an effective defense model.

\vspace{0.5em}\noindent\textbf{Class activation feature based attack:}
Based on above observations, we design a class activation feature based attack (CAFD) method. Our method aims to find strong adversarial examples in class activation feature space by solving the following optimization problem:
\vskip -0.1in
\begin{equation}
\label{eq2}
\begin{aligned}
\max _{\tilde{x}} \Delta(x, \tilde{x})=\delta(\Phi_{x}, \Phi_{\tilde{x}}) \text{, \space \space \space} \\
\text {subject to: }\|x-\tilde{x}\|_{\infty} \leq \epsilon \text{. \space \space \space}
\end{aligned}
\end{equation}

where $\epsilon$ denotes the perturbation budget.
Our attack method is summarized in Algorithm~\ref{alg1}. Given natural examples $x$, we first initialize adversarial examples $\tilde{x}_{0}$ as $x$. Then, we forward $x$ and $\tilde{x}_{t}$ to the pretrained deep neural network $\mathcal{P}$. and obtain their class activation features $\Phi_{x}$ and $\Phi_{\tilde{x}_t}$. Next, we compute the feature distance $\Delta(x, \tilde{x}_t)$ and its gradients using Eq.~\ref{eq2} and Eq.~\ref{eq3}. Finally, we take the gradients to update $\tilde{x}_{t}$ and obtain $\tilde{x}_{t+1}$ using Eq.~\ref{eq4} and Eq.~\ref{eq5}. By iteratively executing such update procedure, Algorithm~\ref{alg1} could maximize $\Delta(x, \tilde{x})$ and output an adversarial example $\tilde{x}$. 

\begin{algorithm}[t]
   \caption{CAFA: Class Activation Feature based Attack}
   \label{alg1}
\begin{algorithmic}[1]
   \REQUIRE A pretrained deep neural network $\mathcal{P}$, natural example $x$, perturbation budget $\epsilon$, number of iterations $T$ and attack step size $\alpha$.
   \ENSURE An adversarial example $\tilde{x}$ with $\|x-\tilde{x}\|_{\infty} \leq \epsilon$.
   \STATE $\tilde{x}_{0} \leftarrow x$;
   \FOR{$t=0$ to $T-1$} 
   \STATE Forward $x$ and $\tilde{x}_t$ to $\mathcal{P}$, and obtain class activation features $\Phi_{x}$ and $\Phi_{\tilde{x}_t}$;
   \STATE Compute the feature distance $\Delta(x, \tilde{x}_t)$ using Eq.~\ref{eq2};
   \STATE Compute gradients $w.r.t$ inputs:
   \vskip -0.15in
   \begin{equation}
   \label{eq3}
    g_{t}=\nabla_{x} \Delta\left(x, \tilde{x}_t\right) \text{;}
   \end{equation}
   \vskip -0.4in
   \STATE Update the adversarial example $\tilde{x}_t$: 
   \vskip -0.15in
   \begin{equation}
   \label{eq4}
    \tilde{x}_{t+1}=\tilde{x}_{t}+\alpha \cdot \operatorname{sign}\left(g_{t}\right) \text{;}
   \end{equation}
    \vskip -0.3in
   \STATE Project $\tilde{x}_{t+1}$ to the vicinity of $x$:
   \vskip -0.15in
   \begin{equation}
   \label{eq5}
   \tilde{x}_{t+1}=\operatorname{clip}\left(\tilde{x}_{t+1}, x-\epsilon, x+\epsilon\right) \text{;}
   \end{equation}
    \vskip -0.3in
   \ENDFOR
   \RETURN $\tilde{x}=\tilde{x}_T$.
\end{algorithmic}
\end{algorithm}

\begin{figure*}[t]
\begin{center}
\centerline{\includegraphics[width= 6.1 in]{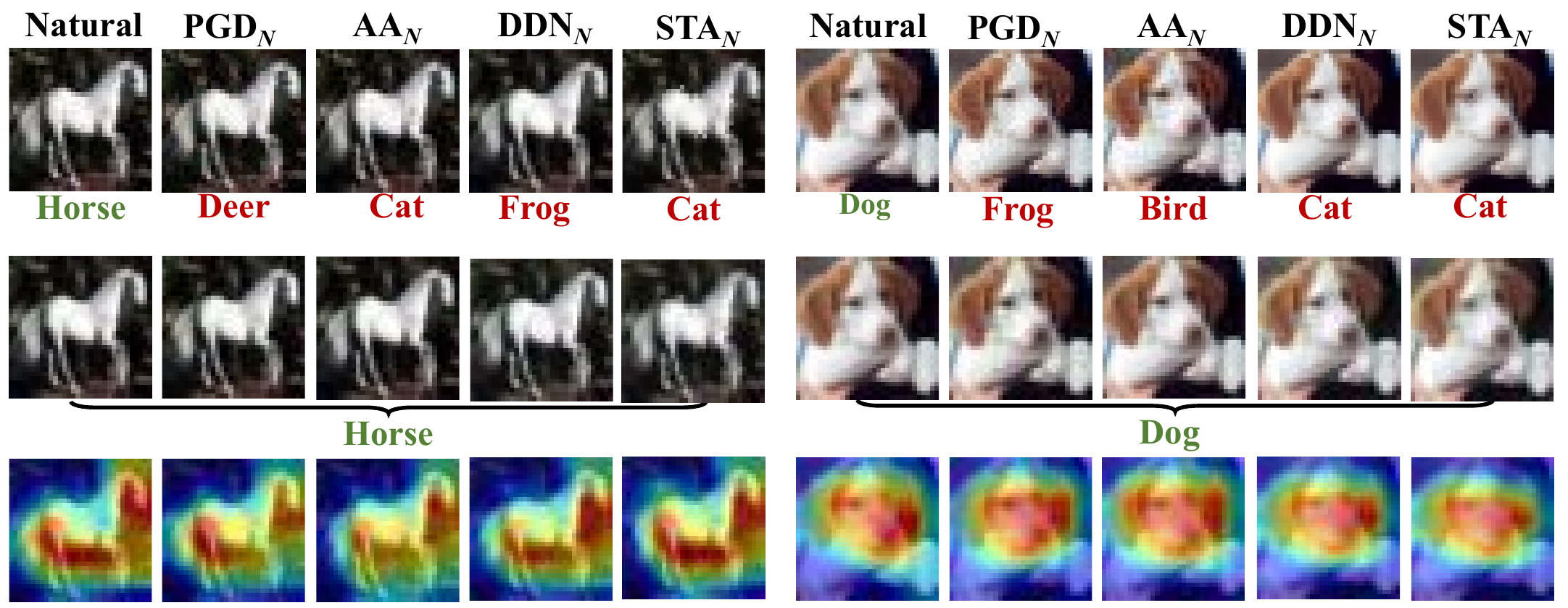}}
\caption{A visual illustration of the defense effect of our model. (\textit{top:} adversarial examples; \textit{middle:} restored examples; \textit{bottom:} class activation maps of restored examples). Subscripts \textit{``N''} indicates that the corresponding attacks are non-targeted attacks.}
\label{fig4}
\end{center}
\vskip -0.4in
\end{figure*}

\subsection{Removing adversarial noise}
\label{section3_3}
We design a class activation feature based denoiser (CAFD) $\mathcal{C}$ to effectively remove adversarial noise. To train the denoiser, we use a hybrid loss function, which consists of a class activation feature loss function and an adversarial loss function.

\begin{algorithm}[t]
   \caption{CAFD: Class Activation Feature based Denoiser}
   \label{alg2}
\begin{algorithmic}[1]
   \REQUIRE Training data $X$, pretrained deep neural network $\mathcal{P}$ and perturbation budget $\epsilon$.
   \REPEAT
   \STATE Simple natural example $x$ from $X$;
   \STATE Craft adversarial example $\tilde{x}$ at the given perturbation budget $\epsilon$ by utilizing Algorithm~\ref{alg1};
   \STATE Forward-pass $\tilde{x}$ through $\mathcal{C}$ and calculate $\mathcal{L}_{caf}$ (Eq.~\ref{eq6} or Eq.~\ref{eq7});
   \STATE Forward-pass $\mathcal{C}(\tilde{x})$ through $\mathcal{D}$, then calculate $\mathcal{L}_{D}$ (Eq.~\ref{eq8}) and $\mathcal{L}_{adv}$ (Eq.~\ref{eq9});
   \STATE Back-pass and update $\mathcal{D}$, $\mathcal{C}$ to minimize $\mathcal{L}_{\mathcal{C}}$ (Eq.~\ref{eq10}) and $\mathcal{L}_{D}$ (Eq.~\ref{eq8});
   \UNTIL $\mathcal{C}$ and $\mathcal{D}$ converge.
\end{algorithmic}
\end{algorithm}

\vspace{0.5em}\noindent\textbf{Class activation feature loss:}
Adversarial examples crafted by CAFA directly disrupt the class activation features and thus lead to misleading predictions. In order to effectively protect target models, the denoiser needs to learn to reduce the distances between natural examples and adversarial examples in the class activation feature space. The class activation feature loss could be defined as follows:
\begin{equation}
\label{eq6}
\mathcal{L}_{caf}=\delta(\Phi_{x}, \Phi_{\mathcal{C}(\tilde{x})}) \text{,}
\end{equation}
where $\Phi_{\mathcal{C}(\tilde{x})}$ denotes the class activation features of the restored example $\mathcal{C}(\tilde{x})$, and $\delta(\cdot)$ denotes the $L_2$-norm distance metric. Considering that $\Phi_{x}$ is the dot product of deep features $F_{x}$ and class weights $W_{x}$, we could also achieve this optimization goal by jointly reducing the distances of deep features and class weights between the natural example and its adversarial example. The class activation feature loss could be modified as follows:
\begin{equation}
\label{eq7}
\mathcal{L}_{caf}=\delta(F_{x}, F_{\mathcal{C}(\tilde{x})})+\delta(W_{x}, W_{\mathcal{C}(\tilde{x})}) \text{,}
\end{equation}
where $F_{\mathcal{C}(\tilde{x})}$ denotes the deep features of $\tilde{x}$ and $W_{\mathcal{C}(\tilde{x})}$ denotes the class weights of $\tilde{x}$. We use Eq.~\ref{eq6} and Eq.~\ref{eq7} respectively to train $\mathcal{C}$ and present their results in Section~\ref{section4.2}. In addition, we empirically observe that removing $\mathcal{L}_{caf}$ loss would result in a significant decrease in the defense effect of the denoiser (See Figure~\ref{fig5}).

\vspace{0.5em}\noindent\textbf{Adversarial loss:}
We introduce an image discriminator $\mathcal{D}$ to enhance the fine texture details of restored examples in the manner of a relativistic average generative adversarial network \cite{jolicoeur2018relativistic}. Compared with a standard generative adversarial network, the relativistic average generative adversarial network is significantly more stable and generate higher quality examples \cite{jolicoeur2018relativistic}. For a given natural example $x$ and its adversarial example $\tilde{x}$ crafted by CAFA, the adversarial loss for $\mathcal{D}$ is defined as follows:
\begin{equation}
\label{eq8}
L_{\mathcal{D}}=-\log \left(\operatorname{sigmoid}\left(\mathcal{D}\left(x\right)-\mathcal{D}\left(\mathcal{C}(\tilde{x})\right)\right)\right) \text{.}
\end{equation}
The adversarial loss for $\mathcal{C}$ is represented by,
\begin{equation}
\label{eq9}
L_{adv}=-\log \left(\operatorname{sigmoid}\left(\mathcal{D}\left(\mathcal{C}(\tilde{x})\right)-\mathcal{D}\left(x\right)\right)\right) \text{.}
\end{equation}
Combining the above class activation feature loss and adversarial loss, the overall loss function for $\mathcal{C}$ is given as:
\begin{equation}
\label{eq10}
\mathcal{L}_{\mathcal{C}}=\lambda_{1} \mathcal{L}_{caf}+\lambda_{2} \mathcal{L}_{adv} \text{,}
\end{equation}

\begin{table*}[t]
\caption{Classification error rates (percentage) of adversarial examples crafted by various of pixel-constrained and spatially-constrained attacks against the VGG-19 target model on \textit{SVHN} and \textit{CIFAR-10} datasets (\textit{lower is better}). CAFD and CAFD$^{'}$ are our defense models corresponding to Eq.~\ref{eq7} and Eq.~\ref{eq6} respectively. The subscript $N$ indicates that the corresponding attack is a non-targeted attack and the subscript $T$ indicates that the corresponding attack is a targeted attack. The compression quality of JPEG is $75$ and the weight of TVM is $0.003$. The DOA method in this paper uses $7 \times 7$ adversarial patches crafted by exhaustive searching to retrain the target model. For each attack, we show the most successful defense with \textbf{bold} and the second one with \uline{underline}.}
\label{tab1}
\vskip 0.2in
\begin{center}
\begin{small}
\begin{tabular}{l|l|ccccccccc}
\toprule [1.0 pt]
\multirow{2}{*}{} &
  \multirow{2}{*}{\textbf{Defenses}} &
  \multicolumn{9}{c}{\textbf{Attacks}} \\ \cline{3-11} 
 &       & None & DDN$_N$  & TI-DIM$_N$       & PGD$_N$   & PGD$_T$       & AA$_N$   & STA$_N$ & STA$_T$ & FWA$_N$ \\ \toprule [1.0 pt]
\multirow{8}{*}{SVHN} &
  DOA    & \textbf{6.37}      & \textbf{7.48}  & \textbf{37.84}      & 38.96  & 32.90       & 39.09  & \textbf{17.03} & \uline{19.10}   & 65.89  \\
 & AT  & \uline{7.16}      & \uline{10.49} & \uline{40.10}      & 31.27 & 25.78      & 32.56 & 20.90 & 21.10   & \textbf{45.26}  \\
 & JPEG  & 9.78      & 11.25 & 92.04      & 95.67 & 86.60      & 97.56 & 37.81 & 33.65   & 93.07  \\
 & TVM  & 10.01      & 21.37 & 88.94      & 94.53 & 75.17      & 95.84 & 27.76 & 23.66   & 96.61  \\
 & APE-G & 10.40      & 11.92  & 89.45      & 93.20  & 83.40       & 81.92  & 43.75 & 42.37   & 86.57  \\
 & HGD & 10.12      & 10.99  & 90.83 & 57.35  & 45.00      & 62.25  & 40.43 & 36.53   & 67.44  \\
 \cline{2-11} 
 & CAFD &
  7.65 &
  14.90 &
  58.57 &
  \textbf {14.64} &
  \textbf{10.63} &
  \textbf{13.57} &
  \uline{19.92} &
  \textbf{18.48} &
  \uline{58.07} \\
 &
  $\text{CAFD}^{'}$ &
  9.79 &
  16.34 &
  61.49 &
  \uline{15.91} &
  \uline{13.46} &
  \uline{14.87} &
  22.19 &
  22.06 &
  60.82 \\ \hline
\multirow{8}{*}{CIFAR-10} &
  DOA    & \uline{7.82}      & 10.42  & 28.53       & 47.81  & 32.37       & 47.43  & \textbf{17.51} & 18.13   & 50.65  \\
 & AT  & 10.34      & 18.07  & 31.78 & 31.77      & 30.17 & 30.61 & 21.16   & 20.69 & 39.70  \\
 & JPEG  & 13.32      & 16.10 & 38.68      & 51.41 & 50.34      & 58.84 & 47.76 & 47.35   & 85.27  \\
 & TVM  & 9.65      & 23.15 & 39.46      & 68.79 & 56.71      & 66.46 & 30.76 & 31.90   & 90.81  \\
 & APE-G & 8.18      & 11.75  & 34.19      & 78.08  & 62.31       & 76.65  & 24.08 & 21.40   & 76.66  \\
 & HGD & \textbf{7.64}      & \textbf{9.18}  & 35.50 & 46.87  & 31.18       & 45.73  & 19.96 & 18.27   & 46.87  \\
 \cline{2-11} 
 & CAFD &
  8.90&
  \uline{9.24} &
  \textbf{26.57} &
  \textbf {12.79} &
  \textbf{10.58} &
  \textbf{11.80} &
  \uline{18.19} &
  \textbf{17.17} &
  \textbf{35.59} \\
 &
  $\text{CAFD}^{'}$ &
  8.95 &
  9.32 &
  \uline{29.85} &
  \uline{15.45} &
  \uline{12.67} &
  \uline{14.27} &
  19.03 &
  \uline {18.10} &
  \uline {39.49} \\ \toprule [1.0 pt]
\end{tabular}
\end{small}
\end{center}
\vskip -0.3in
\end{table*}

where $\lambda_{1}$ and $\lambda_{2}$ are positive parameters to trade off each component. The overall procedure is summarized in Algorithm~\ref{alg2}. Given training data $X$, we first simple natural example $x$ from $X$ and craft its corresponding adversarial example $\tilde{x}$ via CAFA. Then, we forward-pass $\tilde{x}$ through the denoiser $\mathcal{C}$ and calculate $\mathcal{L}_{caf}$. Next, we forward-pass $\mathcal{C}(\tilde{x})$ through $\mathcal{D}$ and then calculate $\mathcal{L}_{D}$ and $\mathcal{L}_{adv}$. Finally, we take a gradient step to update $\mathcal{C}$ and $\mathcal{D}$ to minimize $\mathcal{L}_{C}$ and $\mathcal{L}_{D}$. The above steps are repeated until $\mathcal{C}$ and $\mathcal{D}$ converge.

\section{Experiments}
\label{section4}
In this section, we first introduce the datasets, network architectures and training details used in this paper (Section~\ref{section4.1}). Then, we present and analyze the experimental results of defending against unseen types of attacks and adaptive attacks (Section~\ref{section4.2}). Finally, we conduct an ablation study and an evaluation of the robustness of our model to the perturbation budget, to further show the effectiveness of our defense method (Section~\ref{section4.3}).

\subsection{Experiment setup}
\label{section4.1}

\vspace{0.5em}\noindent\textbf{Datasets:}
We verify the effective of our defense method on two popular benchmark datasets, i.e., \textit{SVHN} \cite{netzer2011reading} and \textit{CIFAR-10} \cite{krizhevsky2009learning}. \textit{SVHN} and \textit{CIFAR-10} both have 10 classes of images, but the former contains 73,257 training images and 26,032 test images, and the latter contains 60,000 training images and 10,000 test images. Images in the two datasets are all regarded as natural examples. Adversarial examples for evaluating defense models are crafted by applying state-of-the-art attacks. These attacks inlcude: (i) pixel-constrained attacks, i.e., PGD \cite{madry2017towards}, CW \cite{carlini2017towards}, AA \cite{croce2020reliable}, DDN \cite{rony2019decoupling} and TI-DIM \cite{dong2019evading,xie2019improving}. (ii) spatially-constrained attacks, i.e., STA \cite{xiao2018spatially} and FWA \cite{wu2020stronger}. Pixel-constrained attacks generally manipulate the pixel values directly by leveraging the $L_p$ norm distance for penalizing adversarial noise, while spatially-constrained attacks focus on mimicking non-suspicious vandalism via spatial transformation and physical modifications \cite{gilmer2018motivating,wu2019defending}.

\vspace{0.5em}\noindent\textbf{Network architectures:}
We use three network architectures to perform classification tasks on \textit{SVHN} and \textit{CIFAR-10}, i.e., a VGG-19 architecture \cite{simonyan2014very}, a ResNet-50 architecture \cite{he2016deep} and a Wide-ResNet architecture \cite{Zagoruyko2016WRN}. The depth and widen factor in the Wide-ResNet architecture are set to $28$ and $20$. The architecture of our denoiser is a DUNET architecture \cite{liao2018defense}. It consists of multiple basic blocks and each block contains a $3 \times 3$ convolutional layer, a batch normalization layer and a rectified linear unit. Our image discriminator is a VGG style discriminator \cite{wang2020basicsr,ledig2017photo}, it consists of a fully connected layer and three convolutional blocks containing convolutional layers followed by a batch normlization layer and a leaky ReLU activation function. 

\vspace{0.5em}\noindent\textbf{Training details:}
For fair comparison, all experiments are conduced on four NVIDIA RTX 2080 GPUs, and all methods are implemented by PyTorch. We use the implementation codes of PGD, DDN, CW and STA methods in the \textit{advertorch toolbox} \cite{ding2019advertorch} and the author's implementation codes of AA, TI-DIM and FWA methods. The default perturbation budget $\epsilon$ is set to $8/255$ for both \textit{SVHN} and \textit{CIFAR-10}. The VGG-19, ResNet-50 and Wide-ResNet networks are used as target models and the VGG-19 network is also utilized as the pretrained network $\mathcal{P}$. Learning rates for target models is $10^{-2}$ on \textit{SVHN} and $10^{-1}$ on \textit{CIFAR-10}. All these networks are pretrained and remain fixed. The denoiser $\mathcal{C}$ and the discriminator $\mathcal{D}$ are optimized using Adam \cite{kingma2014adam}. Their learning rates are initially set to $10^{-3}$ and decay to $2.7\times10^{-5}$ when the training loss converges. The positive parameters $\lambda_{1}$ and $\lambda_{2}$ are set to $10^{2}$ and $5\times10^{-3}$ on \textit{SVHN} and $10^{3}$ and $5\times10^{-3}$ on \textit{CIFAR-10}.

\subsection{Defense Results}
\label{section4.2}

\vspace{0.5em}\noindent\textbf{Defending against unseen types of attacks:}
We use adversarial examples crafted by non-targeted $L_{2}$ norm CW to train previous defense models, and select non-targeted $L_{\infty}$ norm PGD, targeted $L_{\infty}$ norm PGD, non-targeted $L_{2}$ norm DDN, non-targeted $L_{2}$ norm CW, non-targeted AA, non-targeted TI-DIM, non-targeted STA, targeted STA and non-targeted FWA as unseen types of attacks to craft adversarial examples for evaluating defense models. The details of these attacks could be found in appendix A. Figure~\ref{fig4} shows that our method is effective to remove strong adversarial noise. Quantitative analysis in Table~\ref{tab1} demonstrates that our method achieves more robust performance, e.g., reducing the success rate of AA$_N$ from 30.61\% to 11.80\% compared to previous state-of-the-art. The adversarial examples and restored examples are shown in appendix B.

\begin{table}[t]
\caption{Classification error rates (percentage) of attacks against distinct target models with CAFD (\textit{lower is better}). We transfer CAFD, which uses the VGG-19 network as the pretraining network, to the ResNet-50 and Wide-ResNet target models.}
\label{tab2}
\vskip 0.2in
\begin{center}
\begin{small}
\begin{tabular}{l|cllll}
\toprule [1.0 pt]
\multicolumn{1}{c|}{\multirow{3}{*}{Attack}} & \multicolumn{5}{c}{Target Model}                                                \\ \cline{2-6} 
\multicolumn{1}{c|}{} & \multicolumn{1}{c|}{VGG-19} & \multicolumn{2}{c|}{ResNet-50}                            & \multicolumn{2}{c}{Wide-ResNet}                            \\ \cline{2-6}
\multicolumn{1}{c|}{} & \multicolumn{1}{c|}{CAFD}    & \multicolumn{1}{c}{None} & \multicolumn{1}{c|}{CAFD} & \multicolumn{1}{c}{None} & \multicolumn{1}{c}{CAFD} \\ \toprule [1.0 pt]
\multicolumn{6}{c}{SVHN} \\ \toprule [1.0 pt]
PGD$_N$                                         & \multicolumn{1}{c|}{14.64} & 100   & \multicolumn{1}{l|}{11.01} & 97.71   & 21.33 \\
PGD$_T$                                         & \multicolumn{1}{c|}{10.63} & 100   & \multicolumn{1}{l|}{10.62} & 93.85   & 14.14 \\
DDN$_N$                                          & \multicolumn{1}{c|}{14.90} & 99.98 & \multicolumn{1}{l|}{15.53} & 100 & 16.60 \\
AA$_N$                                          & \multicolumn{1}{c|}{13.57} & 100   & \multicolumn{1}{l|}{18.80} & 97.20   & 23.17 \\
STA$_N$                                         & \multicolumn{1}{c|}{19.92} &99.87 & \multicolumn{1}{l|}{22.34} & 96.79 & 23.51 \\
STA$_T$                                         & \multicolumn{1}{c|}{18.48} & 99.71 & \multicolumn{1}{l|}{21.91} & 96.63 & 24.87 \\ \toprule [1.0 pt]
\multicolumn{6}{c}{CIFAR-10} \\ \toprule [1.0 pt]
PGD$_N$                                         & \multicolumn{1}{c|}{12.79} & 100   & \multicolumn{1}{l|}{18.86} & 100   & 20.18 \\
PGD$_T$                                         & \multicolumn{1}{c|}{10.58} & 100   & \multicolumn{1}{l|}{13.19} & 99.91   & 12.450 \\
DDN$_N$                                          & \multicolumn{1}{c|}{9.24} & 99.99 & \multicolumn{1}{l|}{9.25} & 100 & 7.84 \\
AA$_N$                                          & \multicolumn{1}{c|}{11.8} & 100   & \multicolumn{1}{l|}{16.25} & 100   & 17.23 \\
STA$_N$                                         & \multicolumn{1}{c|}{18.19} &100 & \multicolumn{1}{l|}{17.51} & 99.99 & 18.16 \\
STA$_T$                                         & \multicolumn{1}{c|}{17.17} & 99.97 & \multicolumn{1}{l|}{17.07} & 99.66 & 17.93\\ \toprule [1.0 pt]
\end{tabular}
\end{small}
\end{center}
\vskip -0.3in
\end{table}

\begin{table}[t]
\caption{Classification error rates (percentage) under scenarios where defense are leaked. (\textit{lower is better}). Defense models APE-G$^{'}$and HGD$^{'}$ are trained based on adversarial examples crafted by non-targeted PGD with iteration number 20. `It-$\tau$' means that the maximum number of attack iterations is controlled to be $\tau$.}
\label{tab3}
\begin{center}
\begin{small}
\begin{tabular}{lccc}
\toprule [1.0 pt]
\multicolumn{4}{c}{BPDA}                      \\ \toprule [1.0 pt]
\multicolumn{1}{c}{Target} & \multicolumn{1}{c}{Attack} & \multicolumn{1}{c}{Defense} & \multicolumn{1}{c}{Error rate} \\ \hline
APE-G$^{'}$+VGG-19   & PGD$_N$ (It-10)     & APE-G$^{'}$   & 98.32      \\
HGD$^{'}$+VGG-19  & PGD$_N$ (It-10)      & HGD$^{'}$     & 79.50      \\
CAFD+VGG-19  & PGD$_N$ (It-10)      & CAFD    & 47.74      \\
APE-G$^{'}$+VGG-19   & PGD$_N$ (It-20)      & APE-G$^{'}$   & 99.12      \\
HGD$^{'}$+VGG-19  & PGD$_N$ (It-20)     & HGD$^{'}$     & 85.04      \\
CAFD+VGG-19  & PGD$_N$ (It-20)     & CAFD    & 51.25      \\  \toprule [1.0 pt]
\multicolumn{4}{c}{White-box adaptive attack} \\  \toprule [1.0 pt]
\multicolumn{1}{c}{Target} & \multicolumn{1}{c}{Attack} & \multicolumn{1}{c}{Defense} & \multicolumn{1}{c}{Error rate}\\ \hline
APE-G+VGG-19   & DDN$_N$     & APE-G   & 97.85      \\
HGD+VGG-19  & DDN$_N$     & HGD     & 97.10      \\
CAFD+VGG-19  & DDN$_N$     & CAFD    & 93.23      \\
APE-G$^{'}$+VGG-19   & PGD$_N$     & APE-G$^{'}$   & 98.86      \\
HGD$^{'}$+VGG-19  & PGD$_N$     & HGD$^{'}$     & 98.13      \\
CAFD+VGG-19  & PGD$_N$     & CAFD    & 95.18      \\  \toprule [1.0 pt]
\multicolumn{4}{c}{Gray-box adaptive attack}  \\  \toprule [1.0 pt]
\multicolumn{1}{c}{Target} & \multicolumn{1}{c}{Attack} & \multicolumn{1}{c}{Defense} & \multicolumn{1}{c}{Error rate} \\ \hline
APE-G$^{'}$+VGG-19   & DDN$_N$     & APE-G   & 11.07      \\
HGD$^{'}$+VGG-19  & DDN$_N$     & HGD     & 10.93      \\
CAFD$^{'}$+VGG-19  & DDN$_N$     & CAFD    & 10.90      \\
APE-G$^{'}$+VGG-19   & PGD$_N$     & APE-G   & 81.19      \\
HGD$^{'}$+VGG-19  & PGD$_N$     & HGD     & 65.03      \\
CAFD$^{'}$+VGG-19  & PGD$_N$     & CAFD    & 56.76   \\ \toprule [1.0 pt]  
\end{tabular}
\end{small}
\end{center}
\vskip -0.3in
\end{table}

\vspace{0.5em}\noindent\textbf{Cross-model defense results:}
In order to evaluate the cross-model defense capability of our method, we transfer our CAFD model to other classification models, i.e., ResNet-50 and Wide-ResNet. Results in Table~\ref{tab2} present that our method significantly removes adversarial noise crafted by various unseen types of attacks against ResNet-50 and Wide-ResNet. The classification error rates of the ResNet-50 and Wide-ResNet target models are relatively similar to those of the VGG-19 target model, which demonstrates that our method could provide effective cross-model protections. The adversarial examples and restored examples are shown in appendix B.

\begin{figure}[t]
\begin{center}
\centerline{\includegraphics[width=\columnwidth]{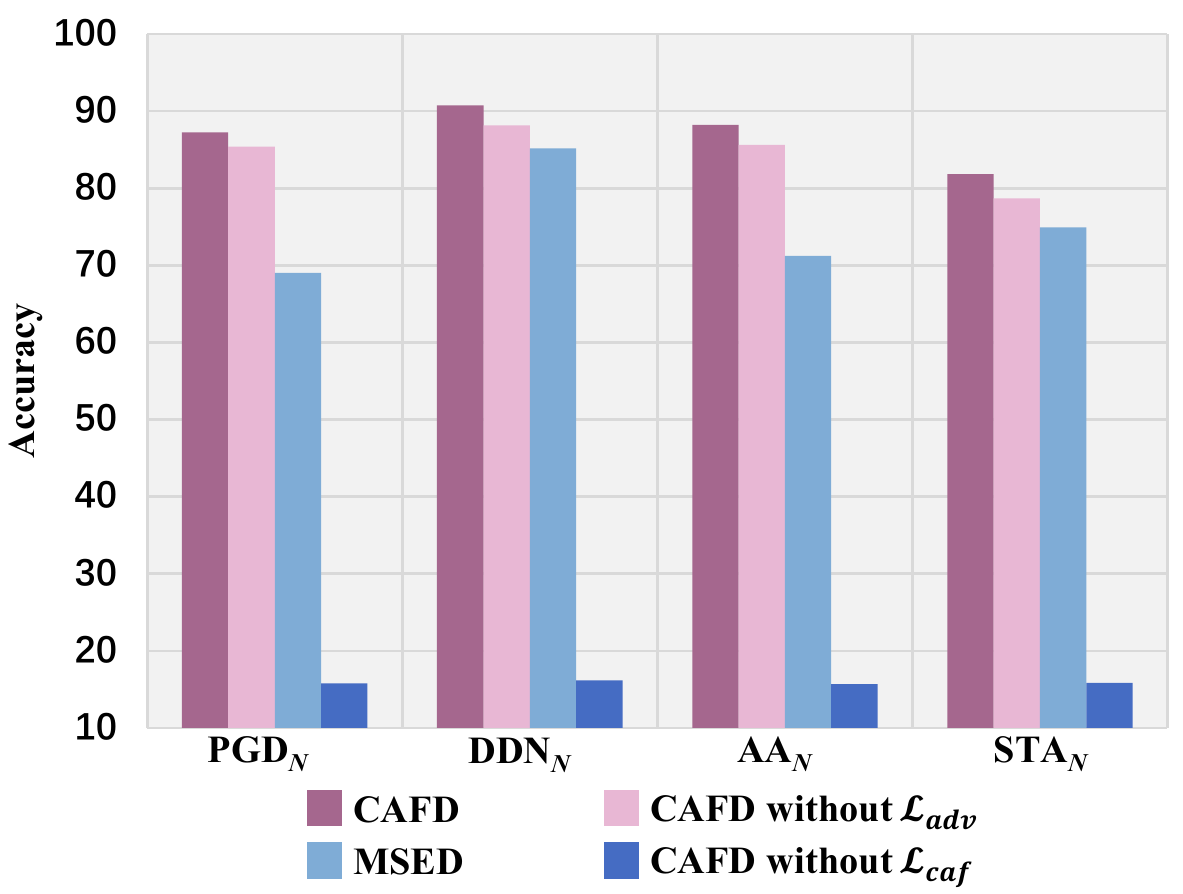}}
\caption{Ablation study on \textit{CIFAR-10}. The figure shows the classification accuracy rates (percentage) of VGG-19 (\textit{higher is better}). The performance of our method without $\mathcal{L}_{caf}$ is significantly affected, which indicates the importance of the class activation feature loss. MSED denotes the defense model which is trained by using a pixel-wise mean square error loss and the adversarial loss.}
\label{fig5}
\end{center}
\vskip -0.4in
\end{figure}

\vspace{0.5em}\noindent\textbf{Defending against adaptive attacks:}
An adaptive attack is one that is constructed after a defense has been specified and been leaked. In this case, the attacker uses the knowledge of the defense and is only restricted by the threat model \cite{athalye2018obfuscated,carlini2017magnet}. We study the following three difficult scenarios: (i) The attacker knows the defense model and uses BPDA \cite{athalye2018obfuscated} to bypass it. (ii) The attacker gains a copy of the defense model and combine it with the original target model into a new target model. Then, the attacker perform a white-box attack on the new target model (iii) The attacker does not directly access the defense model but train a similar local defense model to craft adversarial examples in a gray-box manner. 

In the BPDA scenario, the defense models APE$^{'}$and HGD$^{'}$ are trained based on adversarial examples crafted by non-targeted PGD. We use non-targeted PGD with iteration numbers 10 and 20 to attack defense models in the BPDA manner. As shown in Table~\ref{tab3}, our method shows significant gains, i.e., the classification error rates are reduced by 49.85\% and 39.83\% on average compared to APE and HGD. In the white-box adaptive attack scenario, our method presents a slight reduction in the error rates. Since the defense models are completely leaked to the attacker, defense models' protection capabilities are destroyed under this scenario, which prompts us to strengthen the defense against such attacks in the future. In the gray-box adaptive attack scenario, we use APE-G$^{'}$, HGD$^{'}$ and CAFD$^{'}$ as the local defense models to craft adversarial examples. Our method shows competitive performance against DDN$_N$ and obtains better experimental results against PGD$_N$.

\subsection{Further Evaluations}
\label{section4.3}

\begin{figure}[t]
\begin{center}
\centerline{\includegraphics[width=\columnwidth]{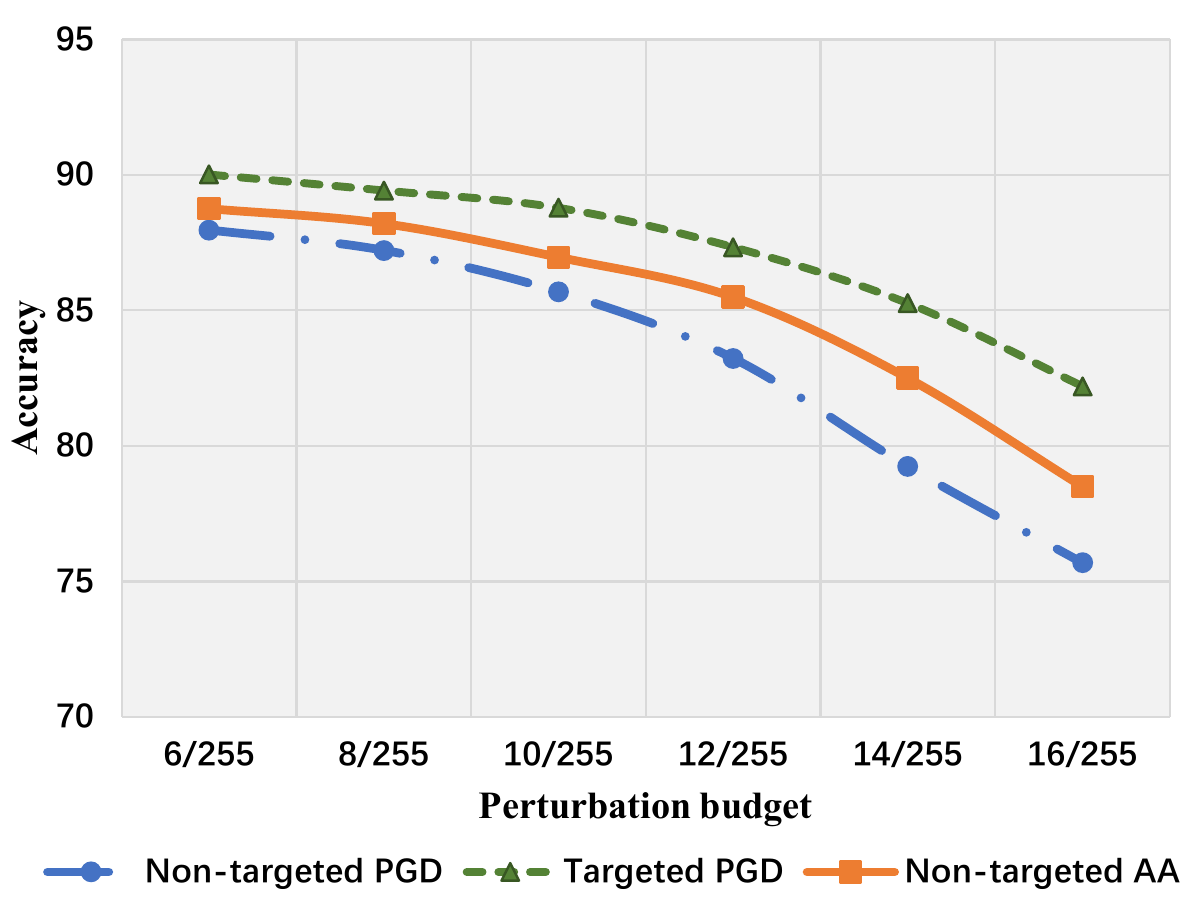}}
\caption{Classification accuracy rates (percentage) of our defense model against adversarial examples with distinct perturbation budget (\textit{higher is better}). We select three strong attacks for this evaluation and set the $L_{\infty}$ norm perturbation budget $\epsilon$ within the range of $(6/255, 16/255)$.}
\label{fig6}
\end{center}
\vskip -0.3in
\end{figure}

\vspace{0.5em}\noindent\textbf{Ablation study:} Figure~\ref{fig5} shows the ablation study on \textit{CIFAR-10}. We respectively remove the adversarial loss $\mathcal{L}_{adv}$ and the class activation feature loss $\mathcal{L}_{caf}$ to investigate their impacts on our model. Removing $\mathcal{L}_{adv}$ slightly reduces the classification accuracy rates because some fine texture details would be lost. The performance of our method without $\mathcal{L}_{caf}$ is significantly affected, which indicated the importance of the class activation feature loss. We train a similar defense model named as MSED by using a pixel-wise mean square error loss and the adversarial loss instead of class activation feature loss. Compared to our defense model, the MSED does not provide sufficient protections against these attacks. 

\vspace{0.5em}\noindent\textbf{Robustness of our model to the perturbation budget:} To explore the robustness of our defense model to the perturbation budget $\epsilon$, we set the $L_{\infty}$ norm perturbation budget $\epsilon$ within the range of $(6/255, 16/255)$ and craft adversarial examples via non-targeted PGD, targeted PGD and non-targeted AA. As shown in Figure~\ref{fig6}, our defense model maintains a relatively high accuracy rate when the adversarial noise is constrained within $(6/255, 12/255)$. This indicates that our model is suggested to defense against attacks with $\epsilon$ less than $12/255$ when the perturbation budget of CAFA is $8/255$. When the perturbation budget continues to increase, the protection effect would reduce significantly. 

\section{Conclusion}
\label{section5}
In this paper, we aim to design a defense model that could mitigate the error amplification effect, especially in the front of unseen types of attacks. Inspired by the observation of the discrepancies between the class activation maps of adversarial and natural examples, we propose a self-supervised adversarial training mechanism to remove adversarial noise in a class activation feature space. Specifically, we first use \textit{class activation feature based attack} to craft adversarial examples. Then, we train a \textit{class activation feature based denoiser} to minimize the distances between the adversarial and natural examples in the class activation feature space for removing adversarial noise. Experimental results demonstrate that our proposed defense model could provide protections against unseen types of attacks and partial adaptive attacks. In future, we can extend this work in the following aspects. First, we need to strengthen the defense against white-box adaptive attacks. Second, we can obtain the class weights for deep features in the internal layers via gradient-based methods, e.g., Grad-CAM \cite{2020Grad} and Grad-CAM++ \cite{chattopadhay2018grad}, so that our method could be applied to more target models or other visual tasks.

\normalem
{\small
\bibliographystyle{ieee_fullname}
\bibliography{egbib}
}

\end{document}